\documentclass[times,twocolumn,final]{article}

\usepackage{amssymb}
\usepackage{latexsym}
\usepackage{natbib}

\usepackage{url}
\usepackage{xcolor}

\usepackage{hyperref}
\usepackage{newtxtext, newtxmath}
\usepackage{subcaption, graphicx}
\usepackage[export]{adjustbox}
\usepackage{svg}
\usepackage{soul}

\usepackage{multirow}
\usepackage{booktabs}
\usepackage{adjustbox}
\usepackage{colortbl}
\usepackage{float}
\usepackage{tikz}
\definecolor{newcolor}{rgb}{.8,.349,.1}

\begin{document}

\title{\textit{Air}: A novel, lightweight Adaptive Image Registration framework}
\author{
  Gabriel De Araujo \\ 
  University of California, Irvine \\ 
  Irvine, CA, USA \\
  \texttt{araujog@uci.edu}
  \and
  Shanlin Sun \\ 
  University of California, Irvine \\ 
  Irvine, CA, USA \\
  \texttt{shanlins@uci.edu}
  \and
  Xiaohui Xie \\ 
  University of California, Irvine \\ 
  Irvine, CA, USA \\
  \texttt{xhx@ics.uci.edu}
}

\maketitle
\begin{abstract}
Image registration has traditionally been done using two distinct approaches: learning based methods, relying on robust deep neural networks, and optimization-based methods, applying complex mathematical transformations to warp images accordingly. Of course, both paradigms offer advantages and disadvantages, and, in this work, we seek to combine their respective strengths into a single streamlined framework, using the outputs of the learning based method as initial parameters for optimization while prioritizing computational power for the image pairs that offer the greatest loss. Our investigations showed improvements of up to 1.6\% in test data, while maintaining the same inference time, and a substantial 1.0\% points performance gain in deformation field smoothness.

\end{abstract}
\section{Introduction}
\label{sec:intro}

A fundamental task for many biomedical imaging applications, image registration has solidified its place in modern medicine. Essentially a problem of transforming (\textit{warping}) one image (labeled as the \textit{moving} image) in order to fit another (the \textit{fixed} image), the most ubiquitous transformation method is known as the deformation field. In order to generate such, there have been two staple paradigms adopted in recent literature on this topic: learning based and optimization based methods.

The learning based methodology commonly consists of training a deep neural network in order to acquire the desired deformation field that will best warp one image to fit its pair. Works like \citep{balakrishnan2019voxelmorph, dalca2018unsupervised,mok2020fast, mok2020large, mok2022affine} show how many different variants to this approach have improved over time, especially since the widespread adoption of transformers-integrated architectures. As such, learning based methods have been adopted at an ever increasing rate within the field.

Optimization based methods had been the traditional approach to image registration prior to the rapid development of deep neural networks, given some of their desirable properties: a smooth deformation field and better numerical stability.

In this work, we sought to combine both of these strategies into one streamlined and novel approach. This architecture, named \textit{\textbf{Air}} (Adaptive image registration), utilizes the predicted deformation field made in-loop by the deep-neural network of choice to serve as the input for the optimization step, where said deformation field is iteratively optimized using the PyTorch \citep{paszke2019pytorch} \textit{Adam} optimizer. 

Furthermore, as a means to preserve performance while reducing the total number of optimization steps, we apply an adaptive method for the selection of image pairs to be further optimized; instead of simply optimizing every single image pair available uniformly, we initialize the optimization step with a small number of optimization iterations for a given pair, and, if it either yields a high enough training loss, or has been randomly chosen for optimization, we increase the enhancement iterations of the respective deformation field. 

The benefits for this are two fold: It allows our framework to perform more optimization steps on pairs that have poorer performance, while also saving computational power on couples with little to no gains when optimized. 

The main contributions of this work are as follows:

\begin{itemize}
    \item We introduce Air, a novel, modular and easy-to-use image registration training framework that incorporates benefits of both optimization and learning based methods into a flexible and streamlined pipeline, capable of being paired with a plethora of different learning based backbones
    \item Air is evaluated using multiple popular state-of-the-art learning-based models, and the atlas-to-patient open-source brain MRI IXI dataset
    \item With no change in inference time, we were able to achieve a 1.6\% improvement in registration performance using a state-of-the-art learning based method as the backbone for our framework, whilst producing a deformation field significantly smoother than the original.
\end{itemize}

\section{Related Work}
\label{sec:related}

\subsection{Optimization based methods}

There are multiple works on addressing the problem of image registration as a mathematical optimization task in the space of displacement vector fields. These works use use a custom function and minimize it for each image pair in iterative fashion. These functions can vary from elastic-type models \citep{bajcsy1989multiresolution}, free-form deformation with B-splines \citep{modat2010fast}, statistic parametric mapping \citep{ashburner2000voxel}, local affine models \citep{hellier2001hierarchical} and Demons  \citep{thirion1998image}. 

Furthermore, the Diffeomorphic approach to image registration, which offers additional desirable traits, mainly topology preservation and transformation invertibility, has been able to achieve similar, and sometimes superior results in various anatomical studies, as seen in its most popular implementations, Large Diffeomorphic Distance Metric Mapping (LDDMM) \citep{beg2005computing}, Symmetric Normalization (SyN) \citep{avants2008symmetric} and DARTEL \citep{ashburner2007fast}. Amid this scope, the deformation is generated by integrating its velocity over time applying the Lagrange transport equation \citep{christensen1996deformable,dupuis1998variational} to achieve a global one-to-one smooth and continuous mapping.
\subsection{Learning based methods}

Learning based Image Registration has recently reached mainstream adoption. VoxelMorph \citep{balakrishnan2018unsupervised} takes advantage of a UNet-like architecture \citep{ronneberger2015u} to regress the deformation fields by minimizing the dissimilarity between a given image pair. Taking this a step further, VoxelMorph-diff \cite{dalca2019unsupervised} introduces the Diffeomorphic registration paradigm to the learning based methodology. Likewise, SYM\_Net \citep{mok2020fast} fuses the symmetric normalization approach from optimization based methods into the learning based approach, simultaneously estimating the forward and backpropagation within the space of diffeomorphic maps.

Recently, more robust transformers-based architectures have sprung into deployment. DTN \citep{zhang2021learning} employs a transformer over the convolutional neural network backbone to capture its contextual relevance and enhance the correspondent extracted features. PVT \citep{wang2021pyramid} addresses major difficulties of implementing transformers in dense prediction tasks with its pyramid-like structure, moving away from the widely used UNet \cite{ronneberger2015u} backbone. TransMorph \cite{chen2022transmorph} introduces a novel registration architecture by using swin transformer blocks as the foundations for its registration framework, and therefore being able to precisely identify spatial equivalences.

This work's proposed  Air framework utilizes the in-loop transitional results from the deep learning based methods discussed above as its optimization based model parameter initialization assumptions, therefore preserving smoothness and maintining topology-preservation characteristics throughout training.

\section{Methodology}

\begin{figure*}
    \centering
    \includegraphics[width=\textwidth]{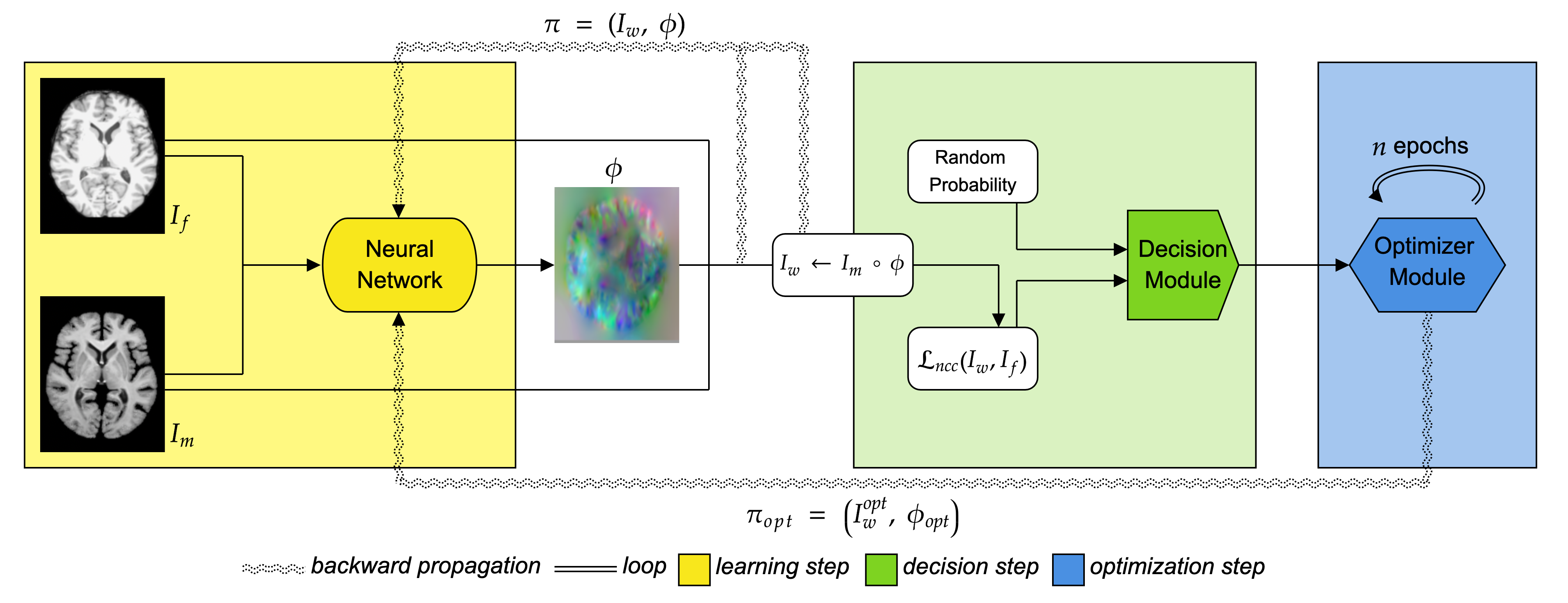}
    \caption{The proposed architecture for our method, Air. Divided into three main sections, Air integrates the stock learning model contained in the learning step and adds two new blocks, the decision and optimization steps. After running the outputs from the neural network into the decision module, if passed through to the optimization step, the optimizer module will loop through $\text{n}_{adp}$ or $\text{n}_{std}$ epochs (5 for $\text{n}_{std}$ and 20 for $\text{n}_{adp}$ in our experiments) and return the optimized output to the neural network for backpropagation calculation of the parameters.}
    \label{diagram}
\end{figure*}

Our proposed novel training architecture, Air, displayed improvement over the state-of-the-art alternatives across the tested dataset for the given medical image registration task. We will describe in greater depth its design and architecture, going over all key details and intuitions of this paradigm.

\subsection{\textbf{\textit{Air}}chitecture}

In order to properly merge both learning and optimization based methods, our proposed method adopts a three step procedure seen in \ref{diagram}, which we will go over in detail over in this section. 

\begin{figure}
    \centering
    \includegraphics[width=0.45\textwidth]{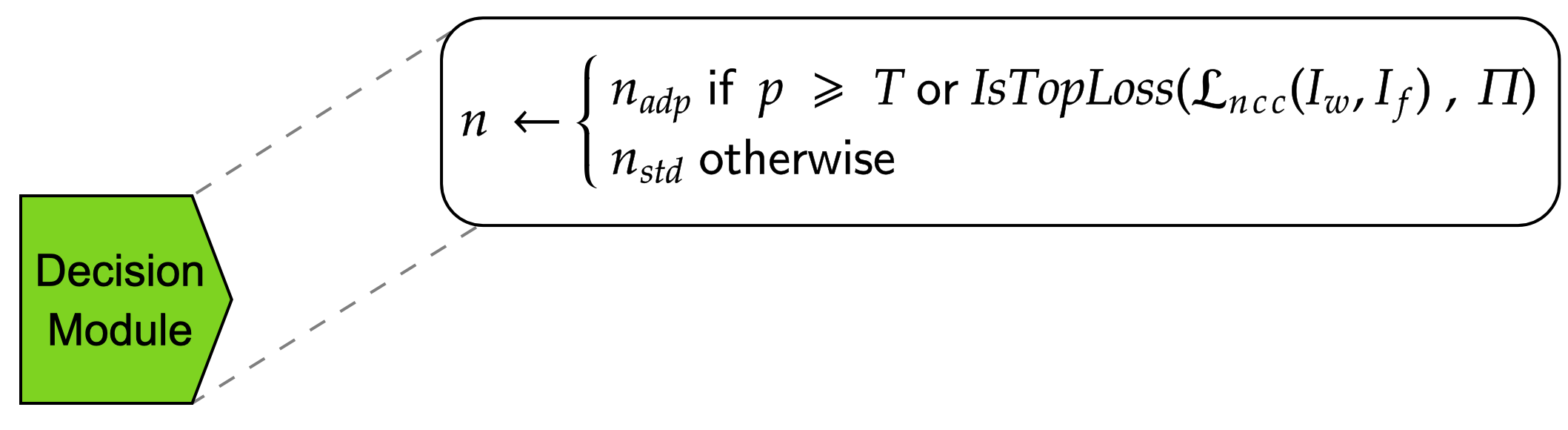}
    \caption{The inner workings of the decision module. $\text{n}_{adp}$ will be assigned if the randomly generated probability $p$ is greater than the current threshold $T$, or the value for $\mathcal{L}_\mathrm{ncc}$ is within the highest in $\Pi$. Otherwise, the standard minimum optimization value $\text{n}_{std}$ will be assigned to $\text{n}$.}
    \label{decisiondiagram}
\end{figure}

\subsubsection{Learning Step}
In this phase, we retrieve the correspondent output for the learning model in a given iteration. For the image registration task, this is the current warped image \textit{$I_w$}, product of the initial moving image \textit{$I_m$} and deformation field \textit{$\phi$} when the following transformation is applied
\begin{align}
    \textit{$I_w$} = \textit{$I_m$} \circ \textit{$\phi$}
\end{align}

\subsubsection{Decision Step}
Afterwards, we enter the decision module. In this phase, as hinted by its name, the architecture will determine, considering the current set $\Pi$ of all examined image pairs in the current epoch $\epsilon$, how many $n$ iterations the given pair $\pi = $ \textit{($I_w$, $\phi$)}, $\pi \in \Pi$, will be optimized. To do so, we established two criteria:

\begin{itemize}
    \item \textbf{Loss function}: Pairs with a high enough loss value when compared to previously examined sets get selected for further optimization. The value is compared only to pairs evaluated in the current epoch, and the threshold is an arbitrary probability set by our team in the training process.
    \item \textbf{Randomness}: In order to combat the issue of convergence in local minima, we integrated a practice similar to simulated annealing into our training method. Specifically, we randomly generate a probability $p$ and attribute a threshold $T$ for random optimization, with an initial value $T_0$, and gradually increase it over the span of training to a set final value $T_\mathrm{final}$, so as to decrease the random nature of further optimizing pairs in $\Pi$. 
\end{itemize}

After concluding this decision process shown in \ref{decisiondiagram}, a value is assigned to $\text{n}$ and the latter is relayed to the framework's optimization pipeline, initiating the next segment of Air's processing. If one of the two criteria above are met, we call $\text{n}$ the \textit{adaptive iteration count}, otherwise it is referred to as the \textit{standard iteration count}.

\subsubsection{Optimization Step}
\begin{figure}
    \centering
    \includegraphics[width=0.41\textwidth]{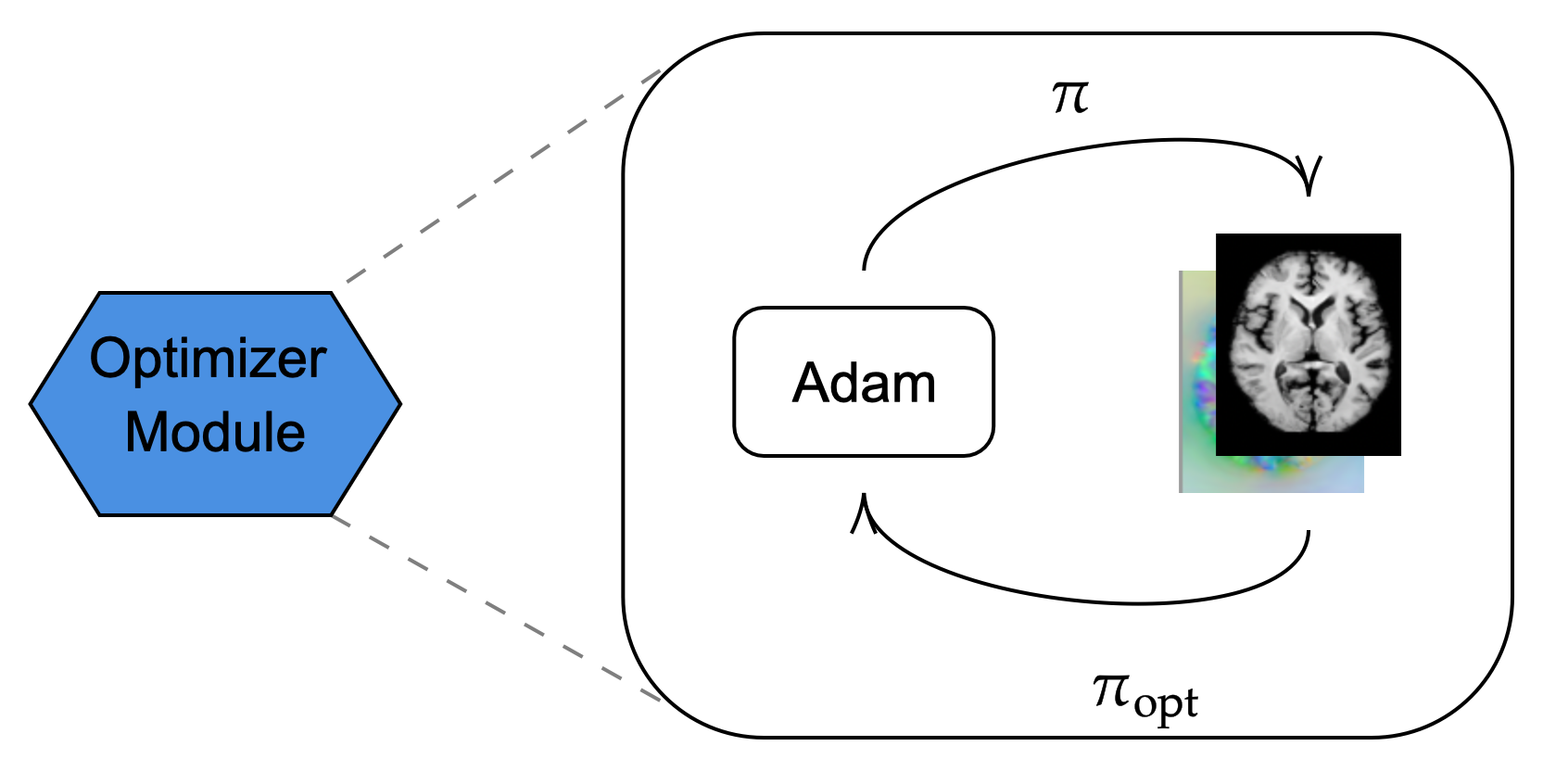}
    \caption{The inner workings of the optimizer module. For $n$ epochs, the Adam stochastic gradient descent optimizer will iteratively optimize the pair $\pi$ and return $\pi_\mathrm{opt}$}
    \label{optimizerdiagram}
\end{figure}

$\pi$ progresses to the optimization block of this architecture in \ref{optimizerdiagram}, where it is iteratively optimized by an optimizer module with the previously set value for n epochs. This produces the new pair $\textbf{$\pi_\mathrm{opt}$} = \textit{($I^\mathrm{opt}_{w}$, $\phi_\mathrm{opt}$)}$, which is subsequently fed to the learning pipeline to perform the loss calculations, backpropagation step and finalize the current iteration.

There are multiple benefits for this structure. It addresses one of the biggest issues of the optimization-based approach of registration, which is parameter initialization. By taking advantage of the processing done by the deep neural network module, we get an improved initial estimate to start the optimization step, reducing computation time considerably whilst taking advantage of a good set of initial parameters for optimization. 

In addition, the use of the \textbf{\textit{decision step}} module means that our framework can perform optimization in adaptive fashion, prioritizing pairs that have obtained poor performance when compared to the rest of the available data $\Pi$ in $\epsilon$, as some of the drawn samples may be close to or have already reached their optimal configuration, and therefore would not benefit much from subsequent optimization.
\subsection{Implementation}
\subsubsection{Optimizer module}
In order to effectively calculate a loss for the data pair $\pi$, we must assign a loss function that takes into account the major elements involved in the learning process: the current warped image $I_w$, the fixed image $I_f$ and deformation field $\phi$. To accomplish this task, we define the local normalized cross correlation ($ncc$) loss between $I_w$ and its ground truth $I_f$ as:
\begin{align}
&{\mathcal{L}_\mathrm{ncc}}(I_w, I_f, p) = \\
\ &\frac{\left[{\sum_{p} (I_w(p) - \bar{I}_w)(I_f(p) - \bar{I}_f)}\right]^2}{\left[\sum_{p} (I_w(p) - \bar{I}_w)\right]^2 \left[\sum_{p} (I_f(p) - \bar{I}_f)\right]^2}
\end{align}
Where $p$ is an element from the set of voxels $\Omega$ of the valuated volume, $\bar{I}_f$ and $\bar{I}_w$ are the respective mean voxel intensities of $I_f$ and $I_w$. This yields our intensity similarity loss function, namely the average negative value of the normalized cross correlation, written as

\begin{align}
    \mathcal{L}_\mathrm{sim} = - \frac{1}{N} \sum_{p \in \Omega} \mathcal{L}_\mathrm{ncc}(I_w, I_f, p)
\end{align}

Where $N$ is the total number of elements in $\Omega$.
Alongside this, we have the diffusion regularization loss $\mathcal{L}_\mathrm{reg}$ to apply smoothness to the deformation field by encouraging the displacement value of a location to be similar to its neighbors

\begin{align}
    \mathcal{L}_\mathrm{reg}(\phi) = \sum_{p \in \Omega } \|\nabla \phi (p) \|_2^2
    \label{regfunction}
\end{align}

Combining both the intensity similarity function $\mathcal{L}_\mathrm{sim}$ and the regularization function $\mathcal{L}_\mathrm{reg}$, we arrive at our assigned loss function for the optimization module $\mathcal{L}_\mathrm{opt}$

\begin{align}
    \mathcal{L}_\mathrm{opt} = \mathcal{L}_\mathrm{sim} + \lambda_\mathrm{opt} \mathcal{L}_\mathrm{reg}
\end{align}

Where $\lambda_\mathrm{opt}$ is the regularization term for the optimization module. This function is responsible for enabling the direct optimization of the deformation field in our pipeline illustrated in \ref{optimizerdiagram}, as it enhances the initially provided $\phi$ in order to minimize the difference between the generated image of the learning step $I_w$ and the target fixed image $I_f$.

\subsubsection{Learning module}

In order to integrate the optimization into the learning process effectively, it is necessary to orient the model to shift its weight calculations towards the optimized parameters in $\pi_{opt}$. To accomplish this, we utilized the Mean Squared Error (MSE) loss function, as it would serve to minimize the difference between the initially generated deformation field $\phi$ and its optimized counterpart $\phi_{opt}$. For this particular problem, we defined MSE as:

\begin{align}
    \mathcal{L}_\mathrm{mse} = \frac{1}{\text{N}_{\phi}} \sum \left[\phi - \phi_\mathrm{opt}\right]^2
\end{align}

Where $\text{N}_{\phi}$ is the total number of parameters in $\phi$. To reach our final learning loss function $\mathcal{L}$, we simply add the regularization function seen in \ref{regfunction} coupled with a regularization term $\lambda$ set for $\mathcal{L}$

\begin{align}
    \mathcal{L} = \mathcal{L}_{\mathrm{mse}} + \lambda \mathcal{L}_\mathrm{reg}
\end{align}

These enable us to adequately exploit the result of the processing done by the stock learning model, apply optimization to its deformation field using $\phi$ as a starting point and successfully applying these enhancements to the learning pipeline, guiding the model to an improved $\phi$ generation performance.

\begin{table*}[htbp]
    \centering
    
    \begin{tabular}{l c c c c}
        \toprule
        Method & $\text{DSC}_\text{val} \uparrow$ & $ \mathbf{|J_{\phi}|}_\text{test} < 0 (\%) \downarrow$ & $\text{DSC}_\text{test} \uparrow$ & $\text{Inference Time}_\text{test} (s) \downarrow$\\
    
        \midrule
        NiftyReg & 0.640 & \textbf{4.270e-5} & 0.637 & 91.42 (CPU)\\
        PVT & 0.726 & 1.760 & 0.728 & 0.273 (GPU)\\
        VoxelMorph-1 & 0.728 & 1.486 & 0.729 &\textbf{0.067} (GPU)\\
        VoxelMorph-1 + Air & 0.738 (+ 1.0\%) & 0.436 (- 1.0\%) & 0.745 (+ 1.6\%) & 0.067 (GPU)\\
        VoxelMorph-2 & 0.735 & 1.420 & 0.732 & 0.096 (GPU)\\
        ViT-V-Net & 0.732 & 1.525 & 0.734 & 0.078 (GPU)\\
        ViT-V-Net + Air & 0.744 (+ 1.2\%) & 0.577 (- 0.94\%) & 0.750 (+ 1.6\%) & 0.078 (GPU)\\
        TransMorph & 0.744 & 1.407 & 0.753 & 0.170 (GPU)\\
        TransMorph + Air & \textbf{0.755 (+ 1.1\%)} & 0.82 (- 0.58\%) & \textbf{0.762 (+ 0.9\%)} & 0.170 (GPU)\\
        \bottomrule
    \end{tabular}
    
    \caption{Comparison of Methods with the Dice Score Coefficient and Negative Jacobian Determinant as primary performance metrics. We can see that Air shows significant improvement in both validation and test data, regardless of the base learning pipeline it is mounted to, while maintaining the same inference time. Alongside this, the percentage of non diffeomorphic units on $\mathbf{|J_{\phi}|}_\text{test} < 0$ is strongly reduced  on all three different learning algorithms, with VoxelMorph displaying an entire percent of lessened sharpness, attesting to the significantly smoother deformation field that is produced from these models.}
    \label{resultstable}
\end{table*}

\section{Experiments}
\subsection{Dataset}

We used a publicly available dataset to evaluate our architecture with atlas-to-patient brain MRI registration. 576 T1–weighted brain MRI images from the Information eXtraction from Images (IXI) was used as the fixed images. The moving image for this task was an atlas brain MRI obtained from \citep{kim2021cyclemorph}. 

We sourced the data with the preprocessed format in \citep{chen2022transmorph}, utilizing a split of 403, 58, and 115 (7:1:2) volumes for training, validation and testing, respectively. The MRI volumes were of size 160 × 192 × 224, and had label maps of 30 anatomical structures to be used in the evaluation process of the registrations.
\subsection{Environment}
All experiments were performed on a NVIDIA RTX A6000 GPU and an Intel Xeon Gold 5218 CPU clocked at 2.30GHz, using PyTorch as the deep learning framework and its Adam optimizer module for both the deep neural network and the optimization step of our architecture. We trained our data for 500 epochs, each having 200 and 20 iterations of randomly paired images of the training and validation sets, respectively, using a batch of size 1, an initial learning rate of 0.0001 and a regularization constant $\lambda$ set to 0.02. 

For our decision module, we used $\text{n}_{adp} = 20$ for the number of \textit{adaptive iteration counts} with a fixed probability of 0.15 of being triggered, and set $\text{n}_{std}$ to 5 on all experiments. For the randomness component, we assigned $T_0 = 0.75$ and $T_\mathrm{final} = 0.95$, and the module to reach $T_\mathrm{final}$ at epoch 450. Our optimizer module has an initial learning rate of 0.1 and a regularization constant $\lambda_\mathrm{opt}$ equal to 1.

\subsection{Evaluation Criteria}

We used the Dice Score Coefficient (DSC) and the percentage of Negative Jacobian Determinant ($\mathbf{|J|} < 0$) from $\phi$ as our primary evaluation metrics. DSC will measure the overlap between the notations of the anatomical structures in the fixed image $I_f$ and the warped image $I_w$, and the non-positive jacobian determinant reflects the degree in which the surrounding space of a voxel is distorted. While a higher DSC is desirable, since it signifies a bigger overlap between \textit{$I_w$} and \textit{$I_f$}, a lower percentage of $\mathbf{|J|} < 0$ points to a more regular structure around a given point in the coordinate grid, and therefore exhibiting a smoother deformation field $\phi$.
\subsection{Baseline}

We compared our proposed architecture with both learning based and optimization based methods, as follows

\begin{itemize}
    \item NiftyReg: We used Sum of Squared Differences (SSD) as the objective function with the default of 300 maximum iterations and a bending energy penalty term of 0.0006
    \item VoxelMorph: We used the default hyperparameter settings for VoxelMorph-1 and VoxelMorph-2 presented in \citep{balakrishnan2019voxelmorph}
    \item ViT-V-Net: We used the default hyperparameter settings for ViT-V-Net in \citep{chen2021vit}
    \item PVT: We used the default hyperparameter settings for PVT in \citep{wang2021pyramid}
    \item TransMorph: We used the default hyperparameter setting for TransMorph in \citep{chen2022transmorph}
\end{itemize}

\section{Results}

\begin{figure*}
    \centering
    \includegraphics[width=0.8\textwidth]{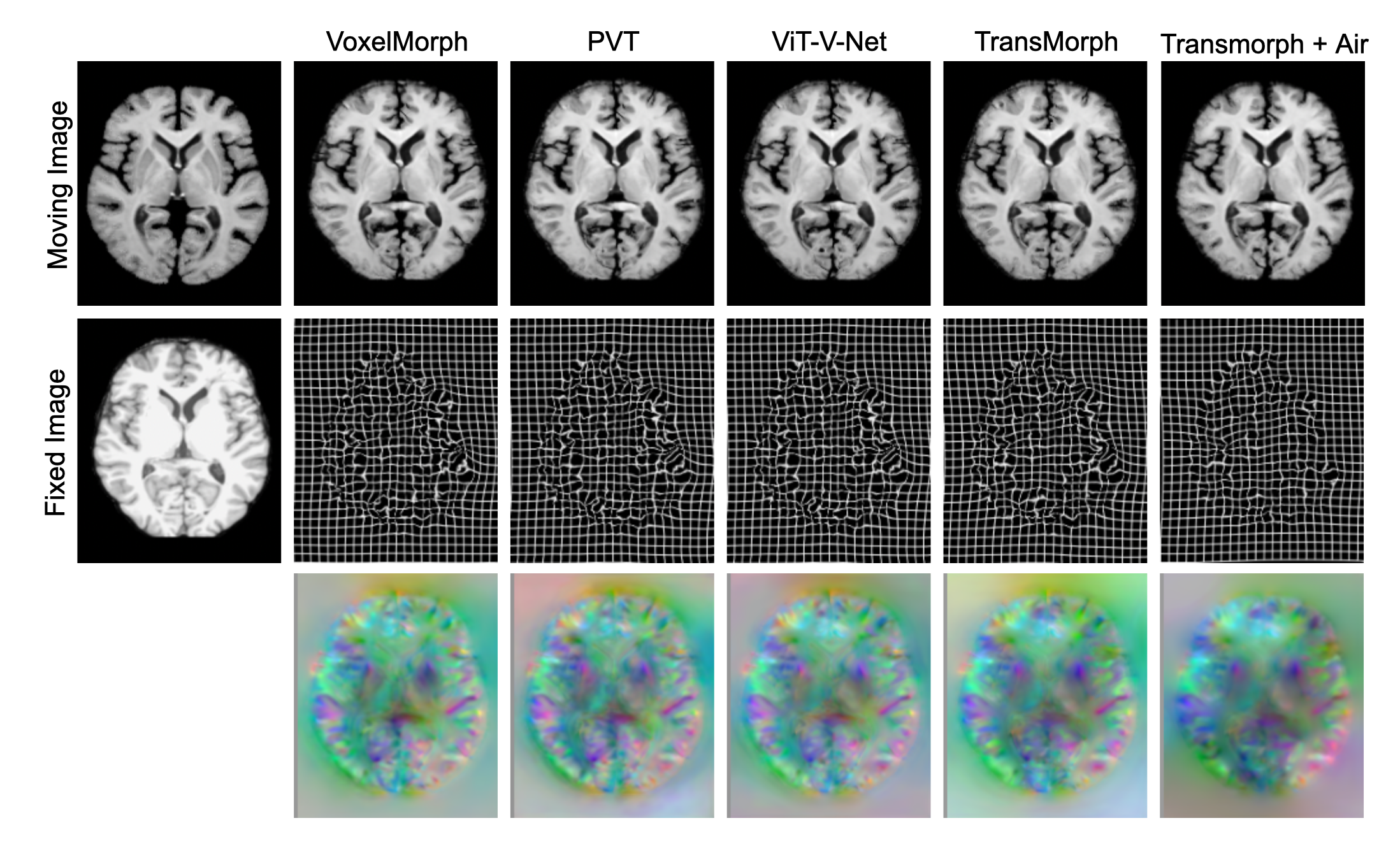}
    \caption{Comparison of model outputs on patient-to-atlas registration on IXI for both deformation grid and field $\phi$ for all learning based methods and Air paired with its highest yielding model for the Dice Score Coefficient, TransMorph.}
    \label{grid+phi}
\end{figure*}

Displayed in \ref{resultstable}, our experiments were deployed integrated with the multiple learning based state-of-the-art models, namely TransMorph, VoxelMorph and ViT-V-Net. Our method employed in this study yielded a considerable improvement of 1.2\% in validation data and a 1.6\% boost in test data evaluation, whilst also providing a reduction of up to 1.0\% in the Negative Jacobian Determinant of $\phi$, attesting to a significant improvement in the smoothness of the outputted deformation field. A general overview of the tested methods can be observed in \ref{grid+phi}, where we can notice the more homogenous nature of the deformation field $\phi$ generated by Air when contrasted with other approaches, like VoxelMorph and PVT.

Air successfully improves performance on both our adopted metrics regardless of its learning based backbone, further corroborating our previous modularity claim. As such, it maintains its usefulness despite advancements in Image Registration paradigms, as it can simply be "plugged in" newer frameworks and boost their performance. 

Though similar, learning-based approaches provide $\mathbf{|J|} < 0$ percentages in their $\phi$ output an order of magnitude above what Air offers, showcasing the clear enhancement in our method. Of course, there is a drastic change in comparison to the optimization based NiftyReg \citep{niftyreg}, as expected, given that smoothness preservation in $\phi$ is one of the key traits in optimization based approaches.

\section{Ablation Study}
\subsection{Loss function}
Originally, our implementation for Air utilized the same loss function for both the optimizer module and learning segment, therefore having the loss function $\mathcal{L}$ as
\begin{align}
    \mathcal{L} = \mathcal{L}_\mathrm{ncc} + \lambda \mathcal{L}_\mathrm{reg}
    \label{wrongloss}
\end{align}
This proved to be ineffective, as we would find it does not provide resources for the model to adjust its parameters to match the optimized deformation field $\phi_\mathrm{opt}$. Such behavior can be explained by the fact that, by utilizing \ref{wrongloss}, while we are feeding the model the optimized version of the initial pair $\pi$, we are not instructing it to modify the parameters so that they are closer to generating $\phi_\mathrm{opt}$, therefore \ref{wrongloss} encourages Air to behave like test-time optimization, effectively ignoring the enhancements done in the optimizer module.
\subsection{Optimizer module tuning}

Another critical aspect of our investigation was the determination of values for both $\text{n}_{adp}$ and $\text{n}_{std}$. As mentioned in our introduction, Air is intended to be a lightweight architecture that cleverly allocates more computational power to instances where a stock learning framework has greater difficulty in establishing parameters for a deformation field grid. Therefore, it is essential to keep $\text{n}_{std}$ to a minimum, while $\text{n}_{adp}$ has to be set where it offers the maximum benefit possible while maintaining computational time within a reasonable bound. 

We evaluated $\text{n}_{std} = 2,5,8,10$ and $\text{n}_{adp} = 10,20,30,50$, and found that, for the former, values lower than 5 seem to be too great of a difference when compared to $\text{n}_{adp}$, which impairs the pipeline's ability to approximate the stock model's output with Air's. Greater values failed to provide additional performance when paired with the different values for $\text{n}_{adp}$. As for the adaptive iteration count, we determined that values higher than 20, while providing a smoother loss curve, significantly hampered the training process, as well as degrade the performance, in similar fashion to what occurred with lower values for $\text{n}_{std}$. While both remaining values showed performance improvements during experimentation, our final choice of 20 displayed a larger gain in both DSC and deformation field smoothness.

\section{Conclusion}
Our investigation introduces a novel training architecture, Air, that combines the strengths of learning and optimization based paradigms, as well as a clever optimization decision algorithm, to produce a powerful, lightweight and efficient framework that allocates computational power to instances where optimization is most needed.

Air displayed an improvement in overall performance compared to the stock training algorithms in both validation and test data, with both DSC and $ \mathbf{|J|} < 0$ metrics supporting this claim throughout different learning modules. Testing in different datasets, like OASIS and MindBoggle, is the natural next step for this project, and is currently being investigated.

\bibliographystyle{model2-names.bst}
\bibliography{refs}

\end{document}